\documentclass[letterpaper, 10 pt, conference]{ieeeconf}  

\IEEEoverridecommandlockouts                              

\usepackage{epsfig} 
\usepackage{mathptmx} 
\usepackage{times} 
\usepackage{amsmath} 
\usepackage{amssymb}  
\usepackage[hidelinks]{hyperref}
\usepackage{url}
\usepackage{siunitx}

\usepackage{censor}
\StopCensoring
\usepackage{adjustbox}  

\usepackage{subcaption}
\usepackage{multirow}
\usepackage{multicol}

\usepackage{booktabs}

\usepackage{graphicx}
\usepackage{algorithm}
\usepackage{algpseudocode}

\newcommand{\orcidicon}{\includegraphics[width=0.32cm]{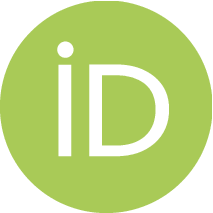}}
\expandafter\xdef\csname orcidCT\endcsname{\noexpand\href{https://orcid.org/\csname orcidauthorCT\endcsname}{\noexpand\orcidicon}}
\expandafter\xdef\csname orcidYG\endcsname{\noexpand\href{https://orcid.org/\csname orcidauthorYG\endcsname}{\noexpand\orcidicon}}

\newcommand{\figscale}{0.95}

\title{\LARGE \bf
SGE: Semantically-Guided Exploration for Unstructured Environments
via Image-Space Waypoint Sampling
}

\author{\censor{Christopher Tatsch$^{1}$\orcidCT{} and Yu Gu$^{1}$\orcidYG{}}
}

\begin{document}

\maketitle
\thispagestyle{empty}
\pagestyle{empty}

\begin{abstract}

This work introduces Semantically-Guided Exploration (SGE), a modular exploration framework for ground vehicles that integrates pixel-level semantic segmentation into sampling-based waypoint selection and receding-horizon route optimization. Unlike conventional geometric exploration methods, SGE evaluates candidate exploration goals directly in the image space using a semantic-aware utility function that accounts for terrain traversability, obstacle proximity, objects of interest, and depth-based exploration reward. Sampled waypoints are projected into 3D and ordered through a real-time Traveling Salesman Problem (TSP) formulation, enabling receding-horizon goal selection. To address real-world navigation uncertainty, the framework introduces mechanisms, including temporary taboo regions to handle navigation failures and a graph-based relocation strategy for efficient backtracking across explored areas. We evaluate SGE in standardized simulation benchmarks against state-of-the-art exploration planners and demonstrate competitive performance in volumetric coverage, while enabling semantic task biasing that cannot be achieved by purely geometric methods. The framework is further validated through real-world experiments using multiple robotic platforms in indoor campus buildings and in limestone and coal mines. Results show consistent performance and adaptability across platforms and domains. 

\end{abstract}

\section{INTRODUCTION}
\label{introduction}

Exploration robots play a critical role across many domains. In space and deep-sea missions, they enable scientific discovery in environments inaccessible to humans \cite{zereik2018challenges, huntsberger2000robotics}. In underground environments and search-and-rescue scenarios following natural disasters, robots can operate where conditions are unsafe or impractical for human responders \cite{kawatsuma2012emergency}. Robots are also increasingly used for environmental monitoring, supporting conservation efforts related to climate change and wildlife protection \cite{dunbabin2012robots}. Across these applications, environments are often hazardous, and communication is unreliable or nonexistent, making teleoperation infeasible and autonomous robotic exploration essential.

\begin{figure}[t]
\centering
\begin{minipage}{\figscale\columnwidth}%
\centering
\begin{minipage}[c]{0.537\linewidth}%
    \centering
    \subfloat[]{%
        \includegraphics[width=\linewidth, trim=390 0 175 0, clip]{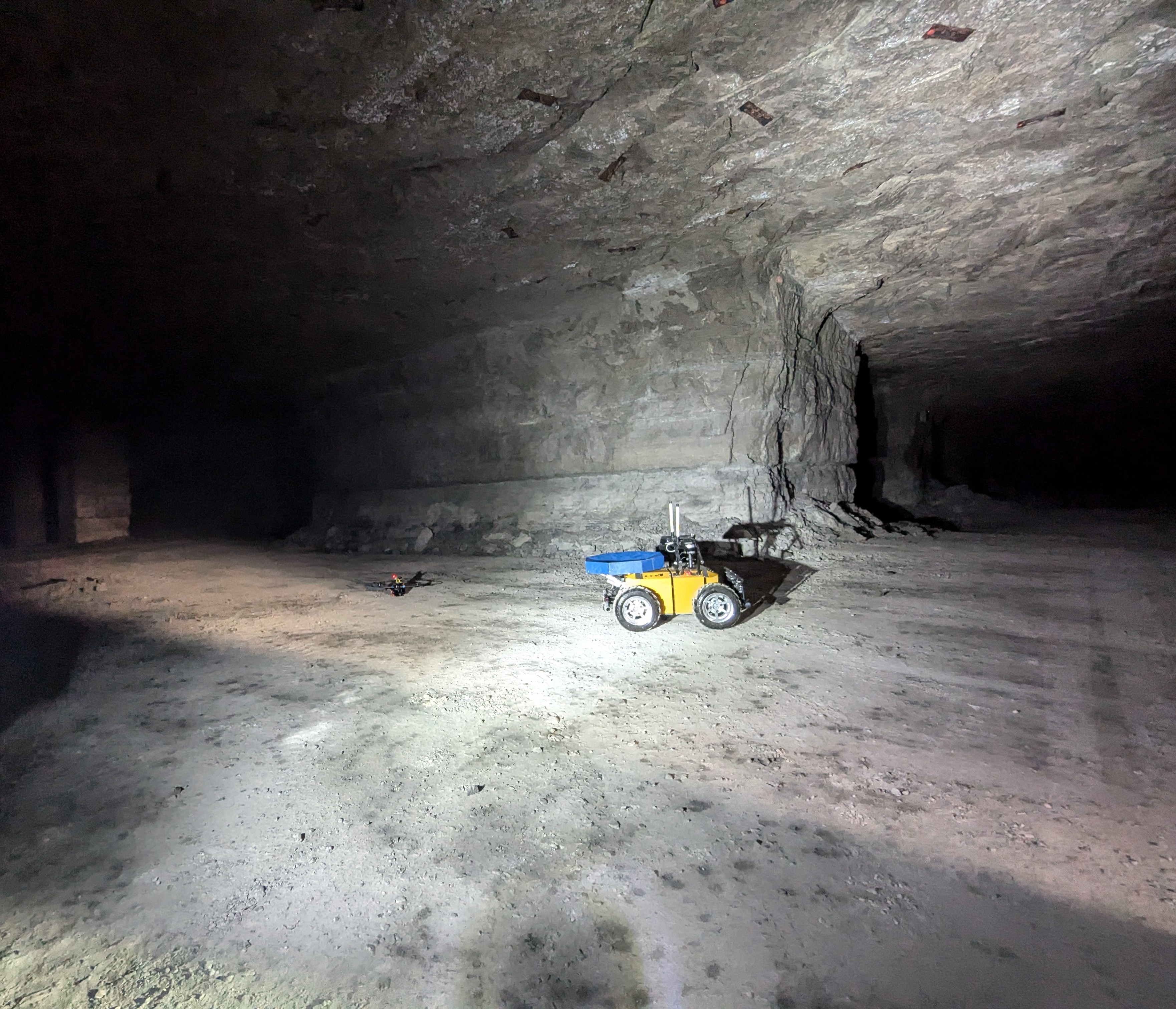}%
        \label{fig:teaser_robot}%
    }%
\end{minipage}%
\hfill%
\begin{minipage}[c]{0.45\linewidth}%
    \centering
    \subfloat[]{%
        \includegraphics[width=\linewidth, trim=0 257 159 48, clip]{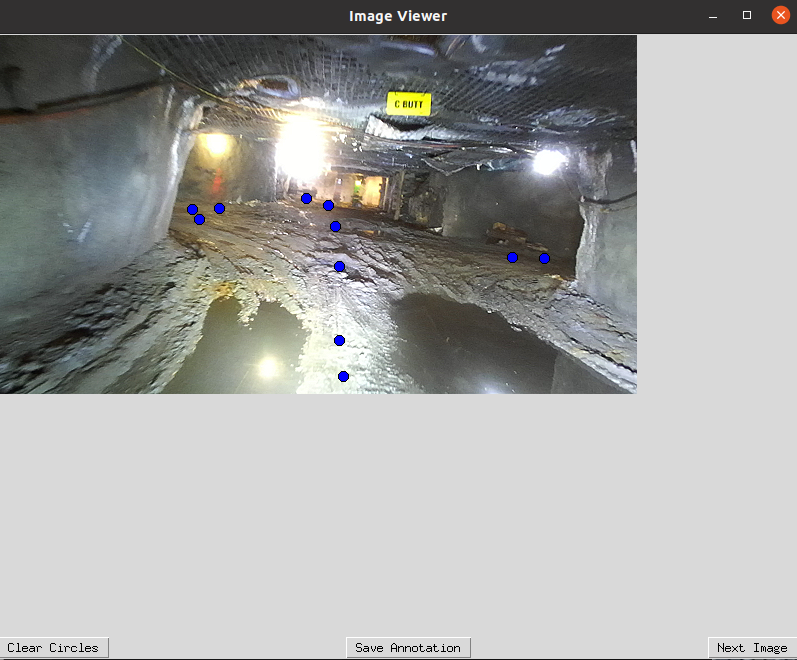}%
        \label{fig:teaser_sampling}%
    }

    \subfloat[]{%
        \includegraphics[width=\linewidth, trim=0 30 0 48, clip]{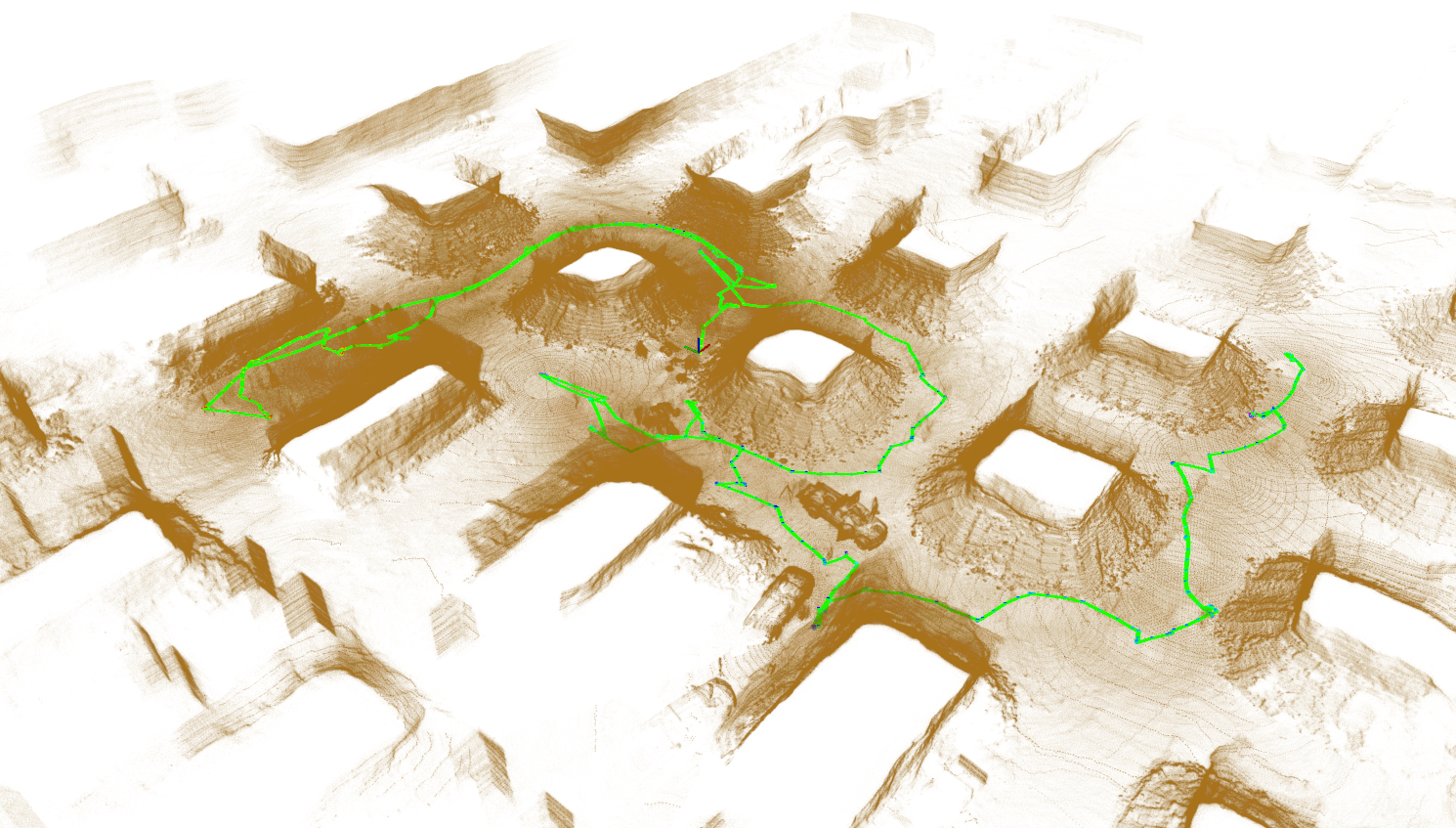}%
        \label{fig:teaser_reconstruction}%
    }%
\end{minipage}%
\end{minipage}
\caption{SGE deployed in underground mines. (a) The \censor{Rhino}
robot exploring a limestone mine. (b) Onboard view during coal-mine
exploration: image-space waypoints (blue) sampled from the semantic
utility distribution avoid regions classified as mud and water.
(c) Resulting 3D reconstruction and executed trajectory (green) in the
limestone mine.}
\label{fig:exploration_example}
\end{figure}

Autonomous robot exploration has advanced significantly in recent years, with systems capable of long-duration operation and accurate mapping in challenging environments, including underground settings \cite{dang2018autonomous}. Nevertheless, most existing exploration algorithms rely primarily on geometric information, such as occupancy, distance, or frontier-based metrics, which are a restricted subset of the rich information available from onboard sensors. Humans, in contrast, leverage prior experience and semantic understanding of a scene when deciding where to go next \cite{sitzmann2018saliency}. Cues such as terrain type, signage, and task-relevant objects are already observable by the robot's camera. This work exploits them directly at the point where exploration goals are generated.

This research aims to enhance autonomous robotic exploration by integrating semantic perception directly into the waypoint selection process of sampling-based planners. We introduce Semantically-Guided Exploration (SGE), a modular framework for unmanned ground vehicles that incorporates pixel-level semantic reasoning into real-time goal generation while preserving the computational efficiency of classical exploration methods. Rather than replacing geometric planning, SGE augments it with semantic-aware importance sampling and receding-horizon route optimization, enabling flexible and task-adaptive behavior across diverse environments. Figure~\ref{fig:exploration_example} shows the robot platform described in \cite{tatsch2023rhino} performing SGE in an underground limestone mine, together with an example of onboard semantic waypoint sampling and exploration environment reconstruction.

The contributions of this work are:
\begin{enumerate}
    \item Semantic-weighted waypoint sampling in image space, where candidate waypoints are generated through a utility-driven formulation that incorporates terrain traversability, obstacle proximity, task-relevant objects, and depth-based exploration incentives. This lightweight formulation enables real-time operation without constructing dense semantic 3D maps.

    \item A receding-horizon TSP-based exploration manager that continuously optimizes candidate waypoints while maintaining scalability for online deployment.

    \item Robust uncertainty-handling mechanisms, including temporary taboo regions for managing navigation failures and a graph-based relocation strategy that enables efficient backtracking across previously explored areas without global replanning.

    \item Extensive validation across simulation and in real-world deployments with two robotic platforms (\censor{Rhino} and Husky ground vehicles) in diverse environments such as indoor campus buildings, limestone and coal mines.
    
\end{enumerate}

The remainder of this paper is organized as follows. Section~\ref{related_work} reviews related work in robotic exploration. Section~\ref{methodology} presents the proposed SGE framework, including semantic-based waypoint selection and the exploration manager. Section~\ref{results} evaluates the approach in simulation and real-world experiments and compares it with state-of-the-art methods. Finally, Section~\ref{conclusion} concludes the paper and outlines directions for future research.

\section{Related Work}
\label{related_work}

Autonomous robot exploration has advanced significantly in recent years. The DARPA Subterranean Challenge (SubT) \cite{orekhov2022darpa} drove many recent advances by requiring ground and aerial robots to autonomously explore challenging underground settings. Similarly, the Autonomous Exploration Development Environment \cite{cao2022autonomous} provides a standardized benchmark for evaluating exploration algorithms under controlled conditions. 

Existing methods can broadly be categorized as information-based, frontier-based, or sampling-based. Information-based methods \cite{bourgault2002information,stachniss2005information} select actions that maximize expected information gain at each robot action; \cite{charrow2015information} and \cite{tabib2016computationally} introduce a Cauchy-Schwarz Quadratic Mutual Information formulation that increases efficiency, enabling higher resolution 3D voxel grids. Frontier-based methods \cite{yamauchi1997frontier,holz2010evaluating,gomez2019topological} drive the robot toward the frontier, which is the boundary between known and unknown space; \cite{kulich2019integrated} samples candidate frontiers and orders them by TSP. Sampling-based planners evaluate subsets of the environment at each time step: NBVP \cite{bircher2016receding} applies a receding horizon over $RRT*$ generated viewpoints, GBPlanner \cite{dang2019graph} uses rapidly exploring random graphs with a global layer for repositioning, and \cite{dharmadhikari2020motion, lindqvist2021exploration, schmid2020efficient} extend the paradigm with motion primitives, model predictive control, and continuous tree rewiring, respectively. Hybrid methods combine paradigms: DSVP \cite{zhu2021dsvp} alternates frontier- and sample-based stages, and TARE \cite{cao2021tare} couples a local viewpoint planner with global TSP optimization and is currently the best-performing method in the Autonomous Exploration Development Environment benchmark. All of these approaches reason over geometric representations derived from LiDAR or occupancy maps. When different terrain types share similar geometry (e.g., mud versus solid ground) or when visual cues such as signage or task-relevant landmarks are informative, purely geometric reasoning may not fully exploit available perceptual information.

Initial efforts to incorporate semantics into exploration differ in where semantics enters the pipeline. A first family builds a semantic 3D map and plans over it. Dang et
al.\ \cite{dang2018visual} annotate an incrementally built map with visual saliency and bias a receding-horizon sampling planner toward salient regions; \cite{asgharivaskasi2023semantic} maintains a semantic octree and derives semantic information gain from them; and \cite{dharmadhikari2023semantics} combines volumetric exploration with inspection or object-centric behaviors around semantic entities. These methods obtain rich global reasoning, but semantics influences planning only after it has been fused into the map, incurring the cost and latency of dense semantic reconstruction. A second family keeps geometric candidate
generation and uses semantics to re-score or search among candidates:
\cite{dang2018autonomous} integrates object detections into an
object-search behavior; \cite{kim2023topological} clusters unexplored
point-cloud regions into a topological graph that guides high-level
decisions; \cite{simons2025segue} scores next-best-view poses by
the semantic features visible from them; and STAGE \cite{patel2024stage}
shifts the endpoint of each local path segment toward the most
traversable pixel of a semantic traversability image, applying
semantics as a local corrective on paths toward geometrically selected
goals. Because candidates are still
generated geometrically, these methods cannot propose goals that only
semantics would distinguish, and effort is spent evaluating candidates
that terrain semantics would have excluded outright.
SGE differs from both in that pixel-level semantics is the primary driver of candidate
generation: waypoints are sampled directly from a semantic utility
distribution over the image plane and only then projected to 3D and
validated geometrically. Semantic influence is applied at bounded
per-frame cost with no dense semantic map, which matters for real-time
operation on embedded platforms; goal generation requires only an
RGB-D camera, and task-conditioning is achieved by adjusting class
weights in the utility function rather than by modifying the planner.



\section{Methodology}
\label{methodology}

\begin{figure*}
\centering
\includegraphics[width=0.90\linewidth]{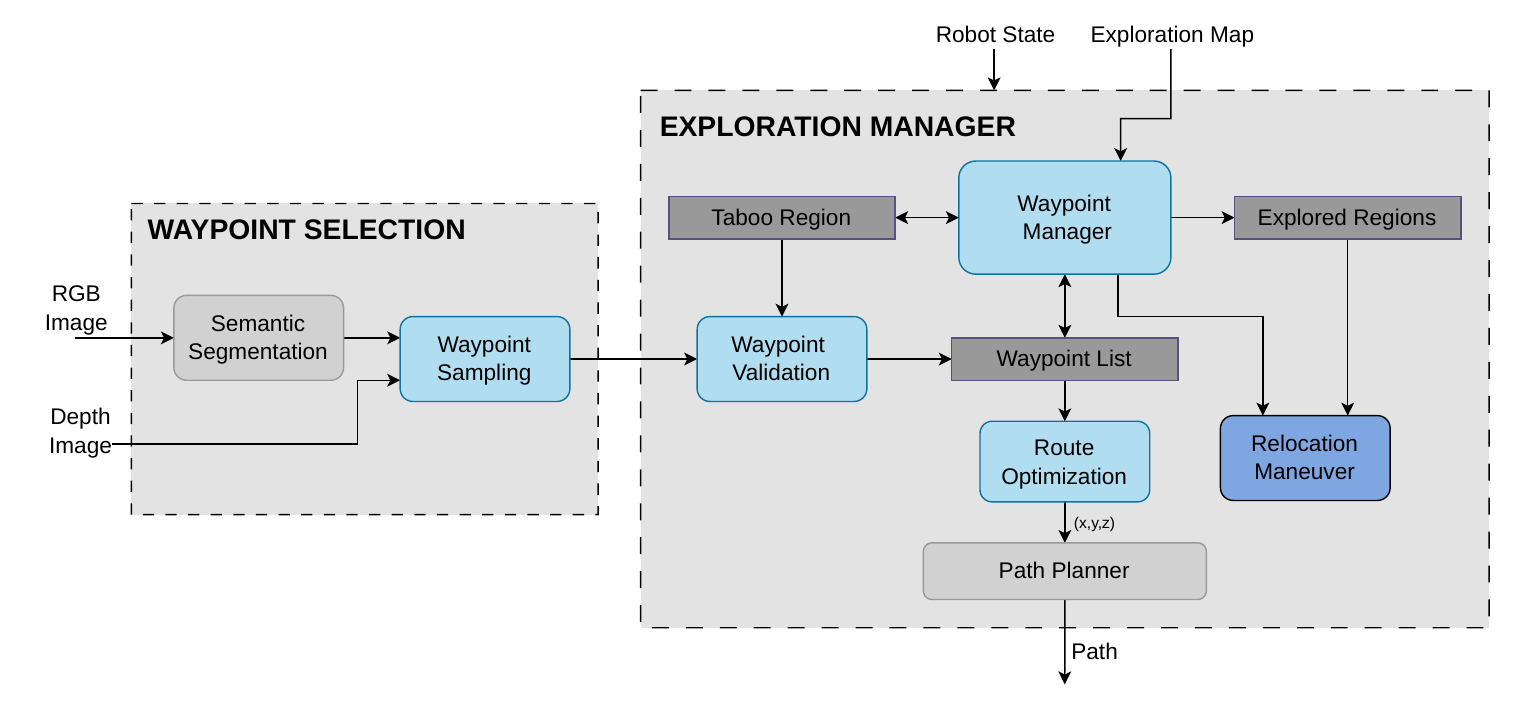}
\caption{Semantic-Guided Exploration (SGE) architecture diagram. Light blue blocks are synchronous modules, dark blue blocks represent asynchronous modules, and light gray blocks correspond to third-party algorithms. Dark gray blocks indicate runtime-evolving data structures whose size changes during exploration.
\label{fig:exploration_overview}}
\end{figure*}

The proposed SGE framework consists of two main components, as illustrated in the architecture diagram in Figure~\ref{fig:exploration_overview}: the Waypoint Selection module, which samples candidate waypoints from semantic image inputs, and the Exploration Manager, which aggregates contextual information and selects the robot’s next goal in a receding-horizon manner. The decision-making process is continuously updated until full environment coverage is achieved. 

The sampling process is guided by semantically segmented images from the robot’s camera, identifying different terrain types and obstacles. Candidate waypoints are ordered by solving a TSP starting from the robot’s current position, and the next waypoint is selected as the first element of the optimized sequence. As new waypoints are generated, the list is continuously updated by adding new candidates and removing visited waypoints. In real-world deployment, exploration is subject to uncertainty: navigation failures may occur, some waypoints may be unreachable due to obstacles, the explored region continuously grows, and the number of candidate exploration goals must remain bounded to ensure real-time TSP optimization \cite{tatsch2020route}. To address these challenges, the framework introduces taboo regions and relocation maneuvers, described in detail in the following sections.

\subsection{Semantically-guided Waypoint Selection}

\begin{figure}[tb]
\centering
\begin{subfigure}[b]{0.32\columnwidth}
    \centering
    \includegraphics[width=\linewidth]{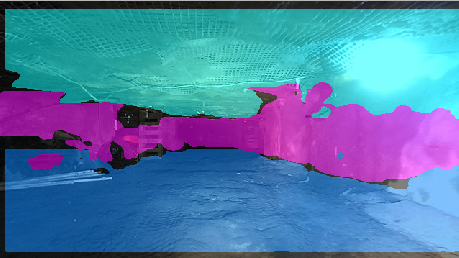}
    \caption{Semantic segmentation}
    \label{fig:viewpoint_sampling_steps_a}
\end{subfigure}
\hfill
\begin{subfigure}[b]{0.32\columnwidth}
    \centering
    \includegraphics[width=\linewidth]{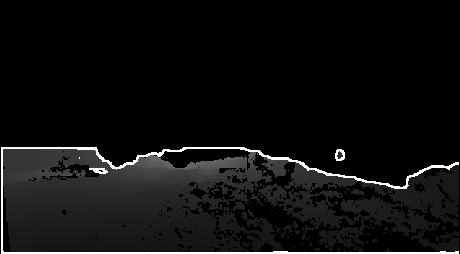}
    \caption{Normalized utility distribution.}
    \label{fig:viewpoint_sampling_steps_b}
\end{subfigure}
\hfill
\begin{subfigure}[b]{0.32\columnwidth}
    \centering
    \includegraphics[width=\linewidth]{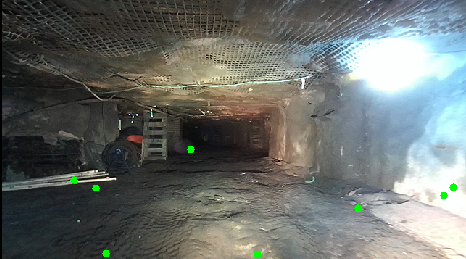}
    \caption{Sampled image waypoints.}
    \label{fig:viewpoint_sampling_steps_c}
\end{subfigure}
\caption{Semantic-guided waypoint sampling process. 
(a) Semantic segmentation of the RGB input. 
(b) Utility-based sampling distribution derived from (\ref{eq:utility_each_point}). 
(c) Candidate waypoints, shown as green dots, selected in the image plane prior to 3D projection.}
\label{fig:viewpoint_sampling_steps}
\end{figure}

The waypoint selection module generates candidate exploration goals directly from semantically segmented RGB images and aligned depth measurements, as shown in Figure~\ref{fig:viewpoint_sampling_steps}. At each time step, the segmentation network produces a pixel-level classification map, from which traversable regions, obstacles, and optional task-relevant objects are identified. Rather than constructing a dense semantic 3D map, waypoint selection is performed directly in the image plane, enabling lightweight real-time operation.

We formulate the waypoint selection process as a semantic-weighted sampling problem defined over the image plane. Rather than uniformly sampling traversable pixels or relying solely on geometric frontier detection, we assign each pixel $i$ a utility score $U(i)$ that reflects its estimated exploration value based on semantic and geometric cues. This formulation enables lightweight evaluation directly in image space, avoiding the need for dense semantic volumetric reconstruction while preserving real-time performance.

For pixels classified as traversable, the utility function is defined as:

\begin{equation}
\label{eq:utility_each_point}
U(i) = \max \left\{ c_a \cdot k_i - \sum_{j=1}^N \left(\frac{c_{bj}}{D_{ij}^2} \right)+ \mathrm{e}^{1-c_c \cdot D_l(i)}+ c_d \cdot z_i, 0 \right\}
\end{equation}

where each term encodes a distinct exploration incentive or penalty. The rectification operator ensures non-negative utilities prior to normalization into a sampling distribution.

The resulting utility values are normalized to form a probability distribution over candidate pixels:

\begin{equation}
\label{eq:probability_sample}
P(i) = \frac{U(i)}{\sum_{m \in I} U(m)}
\end{equation}

where $I$ denotes the set of all pixels in the image. Waypoints are then sampled from $P(i)$, and only the selected pixels are projected into 3D space.

The utility function has four complementary terms:

\begin{itemize}
    \item \textbf{Neighboring Traversable Pixels} ($c_a k_i$): The term $k_i \in [0,8]$ is the count of neighboring pixels classified as traversable. It promotes sampling within traversable regions by favoring pixels surrounded by traversable regions. As a result, it reduces sensitivity to segmentation noise and discourages sampling near semantic boundaries that may correspond to partial obstacles or uncertain classifications. The coefficient $c_a$ controls the relative importance of this spatial consistency term and can be adjusted per terrain type if desired.

    \item \textbf{Distance from Obstacles} ($-\sum_{j=1}^{N} \frac{c_{b_j}}{D_{ij}^2}$): This inverse-squared penalty discourages sampling near obstacle classes. Here, $D_{ij}$ represents the 2D quasi-Euclidean distance from pixel $i$ to obstacle class $j$, and $c_{b_j}$ is the penalty weight for each obstacle type. The inverse-squared formulation ensures strong repulsion near obstacles with smooth spatial decay. Distances are computed using a 2D distance transform inspired by TSDF methods \cite{oleynikova2017voxblox} and implemented via a grassfire algorithm \cite{blum1967transformation}, resulting in $O(V)$ computational complexity, where V is the size of the grid. If an obstacle class is not present in the image, its corresponding distance is set to $\infty$, contributing no penalty.

    \item \textbf{Proximity to Objects of Interest} ($e^{\,1 - c_c D_l(i)}$): This term rewards sampling near task-relevant objects (e.g., directional signs like exit markers), where $D_l(i)$ denotes the 2D quasi-Euclidean distance from pixel $i$ to the nearest object-of-interest class. The exponential form provides stronger emphasis at close range while preserving smooth decay with increasing distance. The coefficient $c_c$ regulates the strength of this semantic bias. If no object of interest is detected, $D_l(i)$ is set to $\infty$, and the term does not influence the utility.

    \item \textbf{Depth-Based Exploration Reward} ($c_d z_i$): The term $z_i$ denotes the depth at traversable pixel $i$. This component favors distant traversable regions, encouraging outward expansion of the explored area. It acts as a surrogate for information gain by promoting waypoints that are likely to reveal new space without explicit volumetric entropy computation. The coefficient $c_d$ determines the influence of depth-based expansion relative to other terms.

\end{itemize}

The weighting coefficients $c_a, c_b, c_c, c_d$ allow task-specific tuning and were selected empirically to balance safety and exploration aggressiveness. An ablation study in Section~\ref{results} quantifies the contribution of each utility term and manager mechanism. To account for segmentation uncertainty, non-traversable but non-obstacle classes receive a small baseline utility, and obstacle pixels zero probability.

After constructing the probability distribution over image pixels, a fixed number of indices $(u,v)$ are sampled. The complexity of this operation is $O(n+mlog(n))$, where $n$ is the number of pixels in the discrete distribution and $m$ is the number of sampled points, which reduces to approximately $O(n)$ in practice, since $m$ is small. Each sampled pixel is projected into the camera coordinate frame using the aligned depth image: $z = z_{depth\_image}$, $x = (u-c_x)\cdot z/f_x$, and $y = (v-c_y)\cdot z/f_y$, where $f_x$ and $f_y$ are the focal lengths and $c_x$ and $c_y$ the principal point coordinates. The resulting 3D points are then transformed into the global frame and passed to the Exploration Manager for validation and optimization.

\subsection{Exploration Manager}

The Exploration Manager coordinates the overall exploration process and consists of two concurrently running loops. The \textbf{optimization loop} selects goals in a receding-horizon manner, validates incoming waypoints, and maintains the optimized goal list. The \textbf{navigation loop} monitors execution, detects navigation failures, updates taboo regions, and triggers relocation behavior when necessary.

The optimization loop is outlined in Algorithm \ref{alg:chapter_03_1stloop}. This algorithm must be fast and scalable to avoid becoming a bottleneck for the navigation loop, where the robot would otherwise have to stop during exploration and wait for new waypoints. The method begins by updating the robot state and sampling new waypoints from the waypoint selection node. Then the Validate Waypoint method validates each new point through a sequence of checks:

\begin{itemize}
    \item Boundary check to ensure the point lies within a predefined exploration region.
    \item Traversability check to verify that the point is within traversable terrain using a local traversability map, a fixed-size region around the robot’s current position. Collision checking here is \( O(n) \) in complexity, where \( n \) represents the number of points in the traversable region map. This check leverages LiDAR data to enhance semantic segmentation, improving the safety of autonomous navigation.
    \item Taboo list check to verify that a waypoint is not inside a list of temporary forbidden locations. This list, although variable in size, is significantly smaller than other data structures. This is implemented by checking whether a sphere of radius $r$ centered at the waypoint intersects any entry in the taboo list.
    \item Exploration region check to confirm that the new point has not yet been explored. Explored regions are represented by a voxel-based structure, where determining if a 3D point falls within a voxel is a constant-time operation, \( O(1) \). While this approach allows efficient checks, it does require fixed-size memory allocation. 
    
\end{itemize}

Valid waypoints are added to a goal list. The next waypoint is selected by continuously solving a TSP with no return, with the robot’s current global position as the starting point. Euclidean distances between waypoints are used to calculate the TSP cost, with the cost matrix updated at each step before optimization. The complexity of evaluating the full cost matrix is \(O(n^2)\), where \(n\) is the number of nodes; however, at each step, only the costs associated with newly added or removed nodes are recalculated. The robot's new distance to all nodes, relative to its current global position, is also computed. 

The default SGE behavior is not to consider the cost of returning to the starting position in the optimization by setting the return-to-depot cost to zero. Alternatively, a return cost to \((0,0,0)\) location can be set, where the robot would consider the cost of returning to the start at each optimization step. The TSP is solved using the Guided Local Search (GLS) metaheuristic \cite{kilby1999guided}, implemented in OR-Tools \cite{or-tools-user-manual}, which provides efficient results with minimal variability in performance and supports real-time receding horizon operation with up to 100 nodes, as shown in \cite{tatsch2020route}.

\begin{algorithm}[t]
\small
\caption{Exploration Manager Optimization Loop}
\label{alg:chapter_03_1stloop}
\begin{algorithmic}[1]

\While{$WaypointList \neq \emptyset$}
    \State $p \leftarrow$ \textproc{UpdateRobotState}()
    \State $w \leftarrow$ \textproc{SampleWaypoint}($p$) 
        \Comment{Semantic-based sampling}
    
    \State $isValid \leftarrow$ 
        \textproc{ValidateWaypoint}($w$, $TraversabilityMap$, 
        $TabooList$, $ExploredRegion$)

    \If{$isValid$}
        \State \textproc{AddWaypoint}($w$)
    \EndIf

    \State $D \gets$ 
        \textproc{UpdateDistanceMatrix}($WaypointList$, $p$)

    \State $OrderedWaypointList \gets$ 
        \textproc{SolveTSP}($WaypointList$, $D$)
\EndWhile
\end{algorithmic}
\end{algorithm}

The Navigation loop is shown in Algorithm \ref{alg:chapter_03_2ndloop}. This loop monitors robot navigation, builds the exploration map, updates taboo regions, triggers relocation behavior, and removes waypoints once explored. Exploration is completed when no valid waypoints remain. This assumes that the robot has a local planner and path follower algorithm capable of reaching goals and maintaining localization. The exploration map is constructed using registered LiDAR point clouds around the robot. At each step, points are marked as occupied in a voxel-based map, which is an $O(n)$ operation, where $n$ is the number of points in the registered scan.

\begin{algorithm}
\caption{Exploration Manager -- Navigation Loop}
\label{alg:chapter_03_2ndloop}
\begin{algorithmic}[1]

\While{$OrderedWaypointList \neq \emptyset$}
    \State \textproc{SendNavigationGoal}($w$)
    \State $p \leftarrow$ \textproc{UpdateRobotState}()
    \State $Exp \leftarrow$ \textproc{UpdateExploredRegion}($p$, $Map$)
    \State $V, E \leftarrow$\textproc{UpdateRelocationManeuverGraph($p$)}
    \If{$\|w - p\| < \tau$ \textbf{or} $expired$}
        \State $w \leftarrow$ \textproc{PopNextWaypoint}($W$)
        \State \textproc{ResetTimer}()
        \State $T_{est} \leftarrow$ \textproc{EstimateTraverseTime}()
        \State $expired \leftarrow \textbf{False}$
    \EndIf
    \For{$w_t \in TabuList$}
        \If{$w_t.timer < 0$}
            \State \textproc{RemoveFromTabuList}($t$)
        \EndIf
    \EndFor
    \If{$T_{nav} > T_{est}$}
        \State $expired \leftarrow \textbf{True}$
        \State \textproc{UpdateTabuList}($w$, $T_{estimated}$)
    \EndIf
    \If{$\|w - p\| > \tau_r$}
        \State \textproc{RelocationManeuver}($w, V, E$)
    \EndIf
\EndWhile
\end{algorithmic}
\end{algorithm}

Taboo regions model temporary navigation failures and uncertainties in the robot's actions. Inspired by Tabu Search metaheuristic \cite{gendreau1994tabu}, they provide memory of temporarily forbidden areas. When a waypoint is sent to the local planner, the Exploration Manager estimates the traversal time from planned path length and robot velocity, and sets a threshold at $150\%$ of that estimate. The manager monitors the navigation at each loop. If the threshold is exceeded, the region around that waypoint is added to the taboo region. This temporary ban lasts $c\times(t_{planned}-t_{planned}')^2$, where $c$ is a constant value, $t_{planned}$ is the estimated time when the waypoint was set, and $t_{planned}'$ is the remaining estimated time. Entries are removed when their timers reach zero, so these regions are only temporarily avoided.

The relocation maneuver is a strategy to safely and efficiently navigate previously explored regions by backtracking along a graph of traversed positions. This process is based on a bidirectional graph \( G = (V, E) \) that represents the regions the robot has already visited, where \( V \) is the set of nodes corresponding to these positions and \( E \) represents edges between these nodes. The graph \( G \) is dynamically updated as the robot moves: if the robot’s current position \( c \) exceeds a threshold distance \( d_{\text{thresh}} \) from any existing node in \( V \), a new node \( v_n \) is added to \( V \), along with edges connecting \( v_n \) to nearby nodes in \( V \) within range \( d_{\text{connect}} \). 

The relocation maneuver is triggered when the goal position \(g \) is far from the robot’s current location \( c \) as determined by the TSP optimization. The first step in this process is to identify the nearest node \( v_s \in V \) to the robot’s current position \( c \) and the nearest node \( v_g \in V \) to the target position \( g \). If \( v_s = v_g \), the goal is already the nearest position from the visited region and the relocation maneuver is not necessary. If \( v_s \neq v_g \), the A* algorithm \cite{hart1968formal} is used to compute the shortest path \( P \) in \( G \) from \( v_s \) to \( v_g \). Euclidean distance is used as the A* heuristic to estimate the cost. Once the path \( P = \{v_s, v_1, v_2, \dots, v_g\} \) is computed, the robot sequentially navigates to each node in \( P \) until it reaches \( v_g \). During the relocation, the Exploration Manager will pause other exploration tasks, waiting for the robot to safely complete the backtracking steps through known terrain.

\section{Experiments and Results}
\label{results}

To evaluate the SGE framework, we conducted simulation benchmarks against state-of-the-art exploration planners and validated the approach in real-world deployments. The experiments were designed to assess (i) volumetric exploration performance, (ii) behavioral characteristics of image-based sampling, and (iii) semantic task-conditioning capability.

Unless otherwise specified, parameters were fixed across experiments to demonstrate robustness across environments. The utility weights were set to \( c_a = 3 \), \( c_b = 5 \), \( c_c = 0.2 \), and \( c_d = 8 \), with eight waypoints sampled per time step. The taboo region was defined as a \(64\,\text{m}^3\) region, and relocation was triggered for goals located more than \(20\,\text{m}\) from the current robot position. These parameters were selected empirically during development to balance safety, exploration aggressiveness, and semantic biasing. The same parameter set was used unchanged across all simulation and real-world experiments, spanning different platforms, sensors, and environments, indicating that SGE does not require per-environment tuning to operate effectively.

\subsection{Simulation Benchmark Evaluation}

SGE was evaluated in the exploration benchmark environment \cite{cao2022autonomous}. Depth data was added to the robot’s camera, and forward-only driving was enforced since waypoint generation depends on camera-aligned perception. Perfect localization was assumed for all planners. The primary evaluation metric was the explored volume. Semantic segmentation used a UPerNet model~\cite{xiao2018unified}  with a ResNet-50
backbone; the framework is agnostic to the segmentation model. The
simulation model was trained on 1500 labeled images from the
benchmark environments~\cite{cao2022autonomous} (80/20 training-validation split), with
eight classes mapped to the utility roles of
(\ref{eq:utility_each_point}) as a per-platform configuration:
traversable (ground, trail, tunnel), small baseline utility (grass,
vegetation), and obstacle (rock, wall, door). All simulations ran on a workstation with an i7-12700K CPU, 64 GB RAM, and an RTX 3090 GPU.

\begin{figure}[t]
\centering
\includegraphics[width=0.90\linewidth]{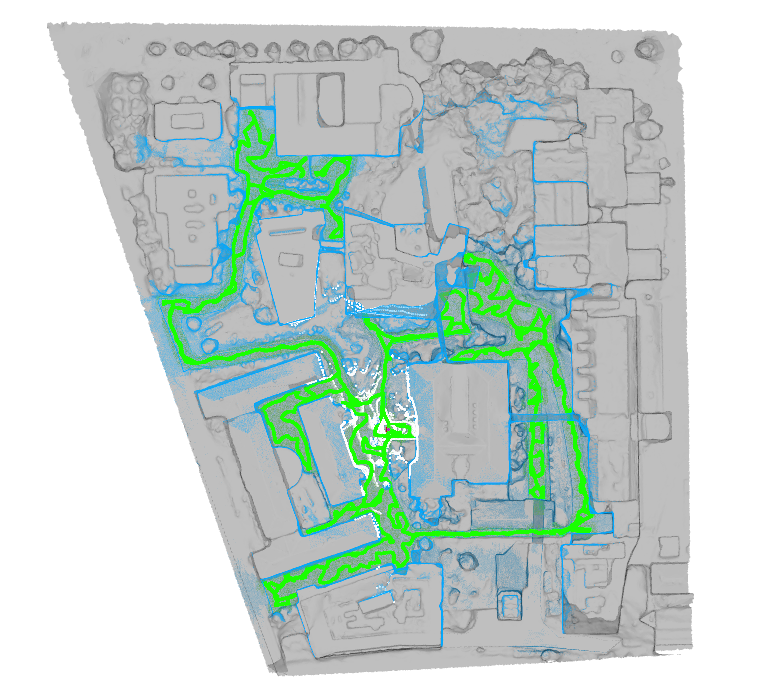}
\caption{Representative exploration trajectory in the campus environment. Green: robot path. Blue: explored volume \label{fig:experiments_01}}
\end{figure}

Figure~\ref{fig:experiments_01} shows a representative exploration run of the SGE method in the campus environment. The robot trajectory is shown in green and the explored volume in blue. The campus layout contains elevation changes, multi-level passages, tunnels, and narrow entrances that are challenging for camera-based exploration. Our method successfully explored all accessible regions, achieving complete environment coverage.

Each method was evaluated five times in the campus environment. A run was considered complete if no further progress was observed for one minute, a one-hour time limit was reached, or the robot returned to its starting position, indicating full coverage. As shown in Table~\ref{tab:campus_environment}, SGE achieved an average explored volume of \(44423\,\text{m}^3\), corresponding to complete exploration of the environment. This performance is comparable to TARE \cite{cao2021tare} and substantially exceeds NBVP \cite{bircher2016receding} in this scenario. While SGE traversed a longer total distance than TARE, this behavior reflects its image-based waypoint generation, which produces shorter-range goals due to camera field-of-view and depth limits; the longer path also yields more fine-grained coverage and consistent exploration progression.

\begin{table}[h!]
\centering
\caption{Exploration Benchmark in Simulation}
\begin{tabular}{c|c|c}
\hline
\multicolumn{3}{c}{Campus Environment} \\ \hline
\textbf{Method} & \textbf{Volume $(m^3)$} & \textbf{Distance $(m)$} \\ \hline
SGE        & 44423                  & 3690                            \\ \hline
NBVP \cite{bircher2016receding}        & 19980                  & 377                           \\ \hline
TARE  \cite{cao2021tare}      & 45875                  & 2341                            \\ \hline
\end{tabular}
\label{tab:campus_environment}
\end{table}

To evaluate the contribution of each SGE component, we conducted an
ablation study following the same benchmark protocol (five runs per
configuration, terminated at one hour or upon completion or failure).
Results are summarized in Table~\ref{tab:ablation_results}. Three
observations emerge.

First, removing any individual component degrades mean explored volume
relative to the full formulation, indicating that each contributes
meaningfully. Removing the depth-based reward ($c_d$) reduced outward
expansion, confirming its role as a surrogate for volumetric
information gain. Eliminating obstacle repulsion ($c_b$) increased
sampling near obstacles and reduced navigation reliability. Removing
the spatial consistency term ($c_a$) increased sensitivity to
segmentation noise, leading to less stable waypoint selection.

Second, ablating the utility terms sharply
increases run-to-run variability. The full formulation achieved
$44423 \pm 659\,\text{m}^3$, whereas ablated variants exhibited
standard deviations up to an order of magnitude larger (e.g.,
$\pm 13743\,\text{m}^3$ without the depth term). The semantic utility
terms therefore contribute not only to mean coverage but to the
reliability and repeatability of exploration: consistent, complete
coverage of the environment was achieved only with the full
formulation.

Third, the Exploration Manager mechanisms provide fault tolerance
under execution uncertainty. Disabling taboo regions or relocation
each resulted in incomplete exploration in 2 of 5 trials. Without
taboo regions, the system repeatedly attempted visually reachable but
physically inaccessible waypoints (e.g., regions at different
elevations or occluded by structural constraints), causing
navigation stalls. Without relocation, the
planner occasionally failed to transition between distant frontiers when the local planner could not complete long traversals. Together, these results indicate a division of roles: the semantic
utility terms govern goal quality and consistency, while the manager
mechanisms safeguard execution under navigation uncertainty.

\begin{table}[t]
\centering
\caption{Ablation Study in Campus Environment (mean $\pm$ std dev over 5 runs)}
\label{tab:ablation_results}
\begin{tabular}{l|c|c|c}
\hline
\textbf{Variant} & \textbf{Volume} (m$^3$) & \textbf{Distance} (m) & \textbf{Runtime} (s) \\
\hline
Full SGE & $ 44423\pm 659$ & $3690 \pm 230$ & $2633 \pm 114$ \\
No Taboo Regions & $42153 \pm 3591$ & $3526 \pm 575$ & $3204  \pm 568$ \\ 
No Relocation & $39216 \pm 6897$ & $3332  \pm 793$ & $2968 \pm 880$ \\ 
\hline\hline
No $c_a$ (Spatial) & $40370\pm 7274$ & $3422 \pm 1336$ & $2768 \pm 982$ \\
No $c_b$ (Obstacle) & $31621 \pm 8408$ & $2199 \pm 915$ & $1922 \pm 834$ \\
No $c_d$ (Depth) & $27160 \pm 13743 $ & $2394 \pm 1706$ & $2151 \pm 1623$ \\ 
\hline
\end{tabular}
\end{table}

Beyond volumetric performance, a key objective of SGE is controllable semantic biasing. To evaluate this capability, we conducted experiments in a modified forest environment \cite{cao2022autonomous} with labeled hiking trails \cite{reed2024autonomous}. Objects of
interest are predefined semantic classes selected per mission,
requiring no runtime annotation. Unlike purely geometric planners,
SGE can bias sampling toward semantically relevant regions by
adjusting the utility weights in (\ref{eq:utility_each_point}): here,
the trail class served as the object of interest and received
slightly higher weights via \(c_a\) and \(c_c\) to favor sampling
along trails.

Because the forest environment is an open-ended coverage problem, exploration time was limited to \(1200\,\text{s}\). This experiment demonstrates semantic adaptability rather than volumetric performance. As shown in Table~\ref{tab:forest_environment}, SGE maintained similar traversal distance while allocating 39.46\% of its trajectory along labeled trail regions. In contrast, TARE, which does not incorporate semantic awareness of trails, traversed only 6.77\% of its path on trail. Although TARE explored slightly greater total volume in the allotted time, SGE demonstrated controllable semantic biasing without sacrificing exploration continuity.

\begin{table}[h!]
\centering
\caption{Exploration Forest Trail Environment}

\begin{tabular}{c|c|c|c}
\hline
\multicolumn{4}{c}{Trail Forest Environment} \\ \hline

\textbf{Method} & \textbf{Volume $(m^3)$} & \textbf{Distance $(m)$} & \textbf{Percentage on trail $(\%)$}\\ \hline
SGE        & 29629                  & 1304    & 39.46                        \\ \hline
TARE        & 39905                  & 1432       & 6.77                     \\ \hline
\end{tabular}
\label{tab:forest_environment}
\end{table}

\subsection{Real-World Exploration}

SGE was validated on two ground robotic platforms across diverse environments, including an indoor campus building, an underground coal mine, and a limestone mine. Both platforms
carried a 32-channel Velodyne Puck LiDAR; the Husky used a ZED2i
stereo camera and \censor{Rhino} an Intel RealSense D435i RGB-D
camera. State estimation and mapping used LIO-SAM
\cite{liosam2020shan}, and navigation the lattice-based FALCO planner
\cite{zhang2020falco}. The field segmentation model was trained on
2500 manually annotated images collected by the robots across the
deployment environments (80/20 split), with ten classes: traversable
(traversable terrain), small baseline utility (rough terrain, grass,
vegetation), and obstacle (mud/water, rock, tree trunk, walls, doors,
fixed structures). The SGE framework operated at the decision-making level, generating and managing exploration waypoints while relying on the underlying navigation stack for execution.

\begin{figure}[t]
\centering
\begin{minipage}{\figscale\columnwidth}%
\centering
\subfloat[Husky robot in \censor{CAC} building.]{%
    \includegraphics[width=0.47\linewidth, height=2.85cm,
        trim=200 0 200 0, clip, keepaspectratio]{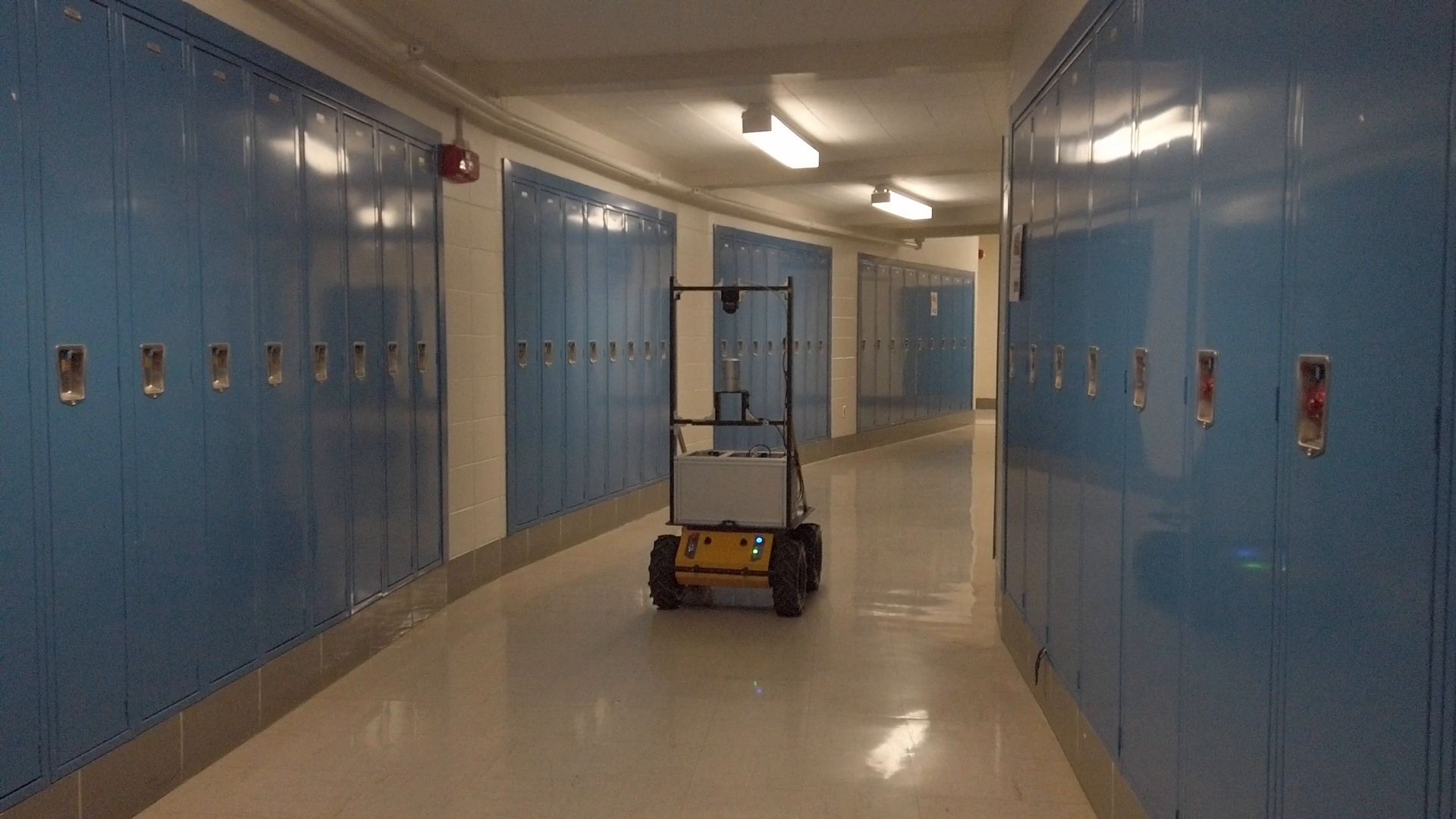}%
}%
\hfill%
\subfloat[Reconstructed map and trajectory \censor{(CAC)}.]{%
    \includegraphics[width=0.47\linewidth, height=2.85cm,
        trim=15 55 25 90, clip, keepaspectratio]{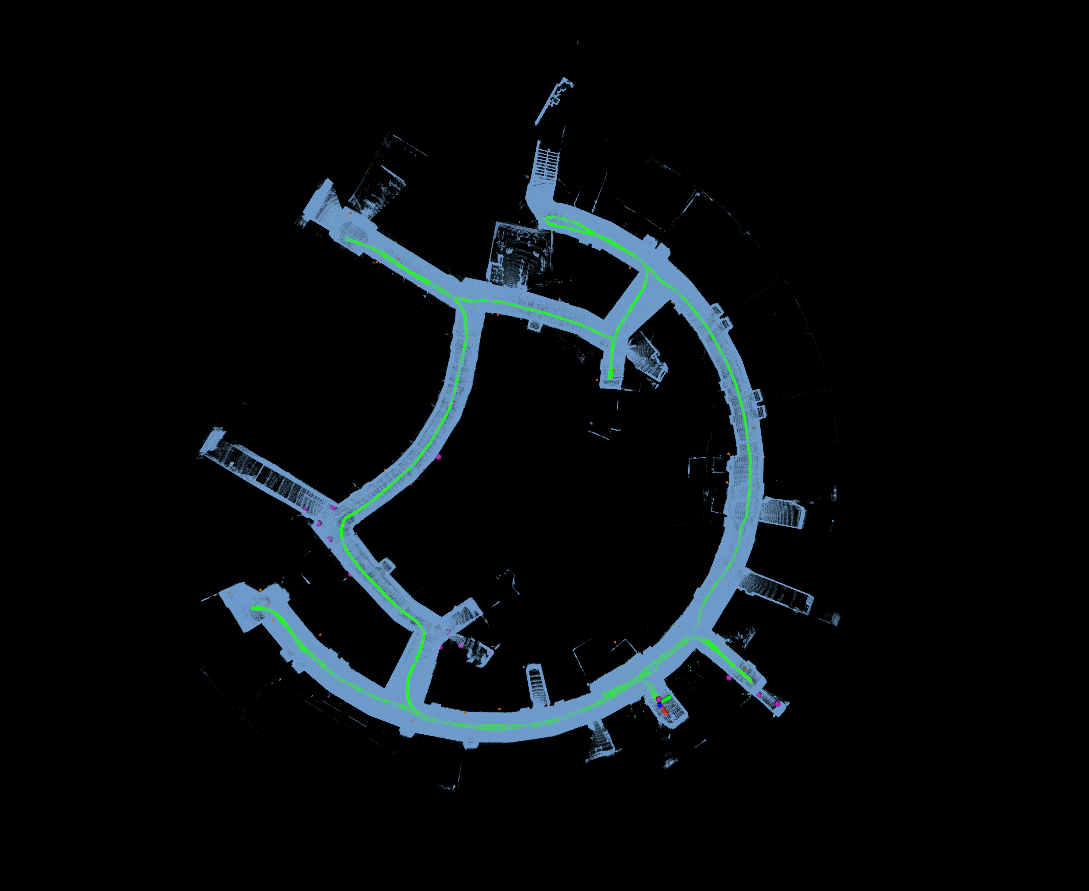}%
}\\[1mm]
\subfloat[\censor{Rhino} deployment in the limestone mine.]{%
    \includegraphics[width=0.47\linewidth, height=2.85cm,
        clip, keepaspectratio]{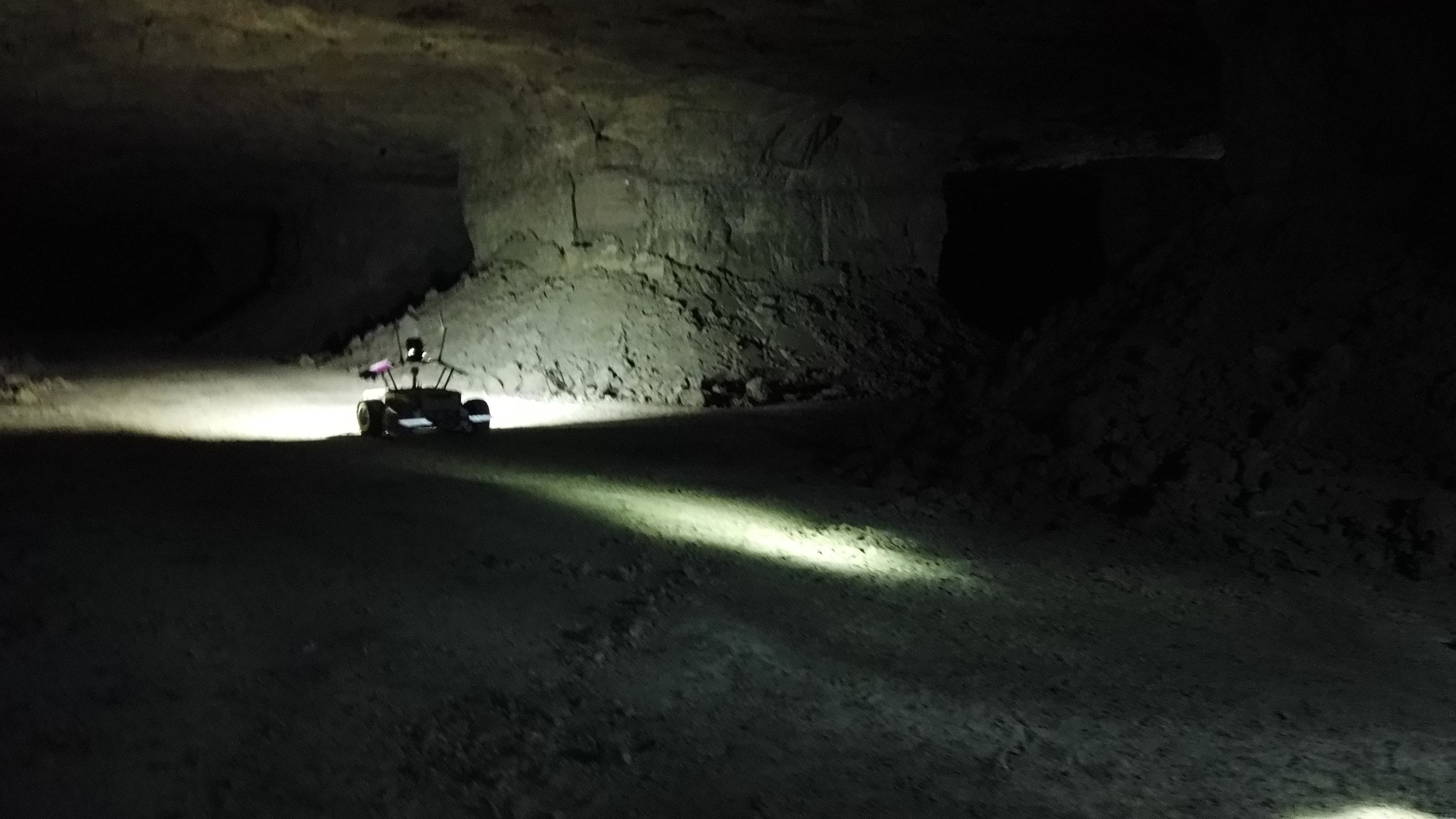}%
}%
\hfill%
\subfloat[Top-down reconstruction and trajectory.]{%
    \includegraphics[width=0.46\linewidth, height=2.85cm,
        trim=52 0 13 0, clip, keepaspectratio]{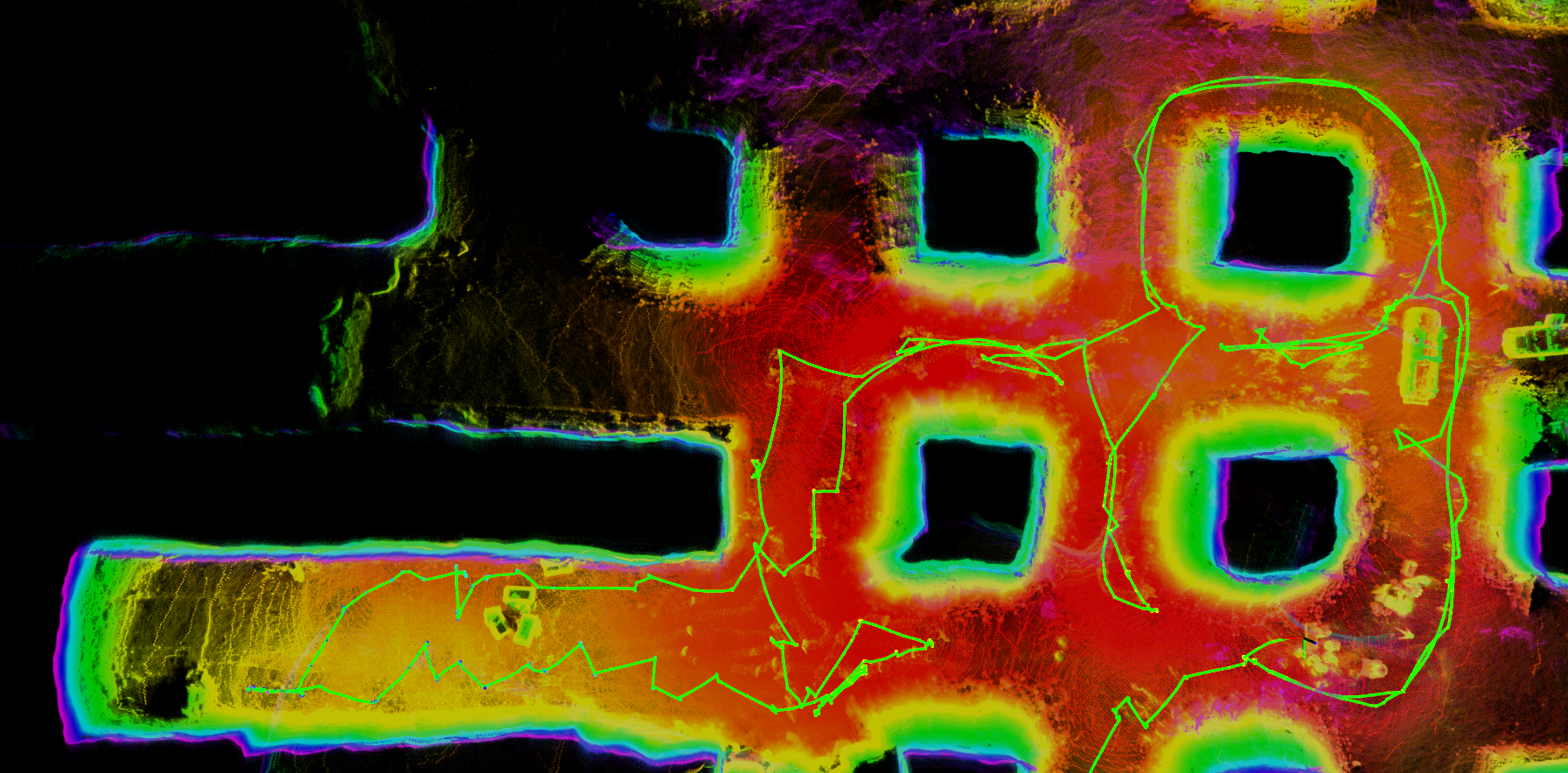}%
}%
\end{minipage}
\caption{Field deployments of SGE.}
\label{fig:sge_examples}
\end{figure}

In the \censor{WVU Creative Arts Center (CAC)}, an indoor building with a non-standard circular layout, intersecting corridors, curved hallways, and pedestrian traffic, SGE successfully explored all accessible areas. The robot traversed $135\,\text{m}$ in $265\,\text{s}$ while executing 35 waypoints. Pedestrians were handled by the local navigation layer without modifying the exploration algorithm. Occasionally, temporarily open classroom doors resulted in waypoints being sampled inside rooms; when such goals proved unreachable, the taboo region mechanism excluded them, allowing exploration to continue. Figure~\ref{fig:sge_examples}(a) shows the Husky robot during exploration, and (b) the reconstructed map with explored volume (blue) and executed trajectory (green).

SGE was deployed in underground coal and limestone mine environments characterized by minimal illumination, dust, and irregular terrain. Figure~\ref{fig:exploration_example}(b) shows the onboard view from \censor{Rhino} in a coal mine, with sampled image-space waypoints in blue. Sampled points avoid regions classified as mud and water, illustrating how semantic utility shapes waypoint selection and improves safety. Two trials were conducted in the \censor{East Fairfield} limestone mine (Figure \ref{fig:sge_examples}(c)), a large-scale environment with open spans up to $12\,\text{m}$ and $9\times9\,\text{m}$ pillar spacing. In the longest run, the robot operated for $2116\,\text{s}$, traversed $765\,\text{m}$, and explored $12835\,\text{m}^3$. Figure~\ref{fig:exploration_example}(c) shows the resulting 3D reconstruction with the executed trajectory in green, highlighting the scale and structural complexity of the environment. A second trial achieved $10240\,\text{m}^3$ over $533\,\text{m}$ in $1462\,\text{s}$, with Figure~\ref{fig:sge_examples}(d) showing the corresponding top-down view. In both trials, exploration ended due to battery and hardware limits rather than algorithmic termination.

Despite segmentation uncertainty in low-visibility conditions, the Exploration Manager maintained stable behavior. Misclassified regions occasionally produced unreachable waypoints (e.g., near pillars), which the taboo mechanism excluded, preventing repeated attempts and allowing exploration to proceed. When the TSP optimizer selected distant frontiers, relocation maneuvers enabled structured backtracking through explored regions. These behaviors were consistent across deployments, indicating resilience to perception noise and navigation uncertainty.

\section{Conclusion and Future Work}
\label{conclusion}

This work presented SGE, an exploration framework in which pixel-level
semantics is the primary driver of waypoint generation. Across
simulation benchmarks and field deployments, SGE achieved reliable
large-scale exploration in real time, with volumetric coverage
comparable to state-of-the-art geometric planners while additionally
providing task-conditioned behavior that purely geometric methods do
not, particularly in structurally complex environments.


Semantic-guided exploration represents an initial step toward fully learning-based exploration. Future work will investigate online terrain-aware cost adaptation, train visual attention–based networks for direct waypoint sampling from RGB-D inputs, and conduct broader evaluations in extreme terrains where hazardous and unpredictable conditions rigorously test robustness.








\bibliographystyle{IEEEtran}
\bibliography{bibliography}

\end{document}